\documentclass{article}

\PassOptionsToPackage{numbers, compress}{natbib}

\usepackage[final]{neurips_2020}

\usepackage[utf8]{inputenc} 
\usepackage[T1]{fontenc}    
\usepackage{hyperref}       
\usepackage{url}            
\usepackage{booktabs}       
\usepackage{amsfonts}       
\usepackage{nicefrac}       
\usepackage{microtype}      
\usepackage{multirow}
\usepackage{graphicx}
\usepackage{bm}

\title{Real-time parameter inference in reduced-order flame models with heteroscedastic Bayesian neural network ensembles}

\author{%
  Ushnish Sengupta \thanks{The first two authors contributed equally to this work.} \\
  Department of Engineering\\
  University of Cambridge\\
  Cambridge, UK \\
  \texttt{us271@cam.ac.uk} \\
  \And
  Maximilian L. Croci {\footnotemark[1]}\\
  Department of Engineering \\
  University of Cambridge\\
  Cambridge, UK\\
  \texttt{mlc70@cam.ac.uk} \\
  \And
  Matthew P. Juniper\\
  Department of Engineering \\
  University of Cambridge\\
  Cambridge, UK\\
  \texttt{mpj1001@cam.ac.uk} \\
}

\begin{document}

\maketitle

\begin{abstract}
The estimation of model parameters with uncertainties from observed data is an ubiquitous inverse problem in science and engineering. In this paper, we suggest an inexpensive and easy to implement parameter estimation technique that uses a heteroscedastic Bayesian Neural Network trained using anchored ensembling. The heteroscedastic aleatoric error of the network models the irreducible uncertainty due to parameter degeneracies in our inverse problem, while the epistemic uncertainty of the Bayesian model captures uncertainties which may arise from an input observation's out-of-distribution nature. We use this tool to perform real-time parameter inference in a 6 parameter G-equation model of a ducted, premixed flame from observations of acoustically excited flames. We train our networks on a library of 2.1 million simulated flame videos. Results on the test dataset of simulated flames show that the network recovers flame model parameters, with the correlation coefficient between predicted and true parameters ranging from $0.97$ to $0.99$, and well-calibrated uncertainty estimates. The trained neural networks are then used to infer model parameters from real videos of a premixed Bunsen flame captured using a high-speed camera in our lab. Re-simulation using inferred parameters shows excellent agreement between the real and simulated flames. Compared to Ensemble Kalman Filter-based tools that have been proposed for this problem in the combustion literature, our neural network ensemble achieves better data-efficiency and our sub-millisecond inference times represent a savings on computational costs by several orders of magnitude. This allows us to calibrate our reduced order flame model in real-time and predict the thermoacoustic instability behaviour of the flame more accurately.
\end{abstract}

\section{Introduction}

Complex, nonlinear physical models of engineering systems give rise to challenging inverse problems which require the estimation of unknown model parameters from observations, along with uncertainty quantification. With the growing popularity of digital twins \citep{9103025}, there is also a need to do bring down the computational cost of parameter inference such that these models may be updated in real-time using the latest sensor observations. Traditional filtering-based techniques require the system to be simulated in parallel, which becomes infeasible if the underlying model is computationally expensive to simulate. Amortized inference using neural networks \citep{Cranmer201912789} circumvents this problem by having an expensive offline training phase, where a surrogate of the approximate posterior $p(\boldsymbol{\theta} | \mathbf{z})$ is learnt from a library of simulator-generated observations $\mathbf{z}_i$ corresponding to input parameters $\boldsymbol{\theta}_i$, typically using normalizing flow-based methods \citep{rezende2015variational} \citep{radev2020bayesflow}. The surrogate can then be rapidly evaluated online to perform parameter inference on new observed data.

The physical model that motivated this paper is a reduced order kinematic model for flames that captures the response of a ducted premixed flame to acoustic perturbations, known as the G-equation\cite{dowling1997nonlinear}. An accurate description of this response is necessary because the interplay between acoustic waves and the heat release rate of the flame can lead to undesirable thermoacoustic instabilities, which have presented an obstinate challenge to designers of high-energy density combustors \cite{lieuwen2003modeling}. Previous studies \cite{yu2019data}\cite{yu2019time} have reported that calibrating the parameters of the G-equation model parameters using the Ensemble Kalman Filter can make it quantitatively accurate and reproduce the experimentally observed behaviour of acoustically forced flames. To achieve a converged estimate of the parameters, however, the ensemble Kalman filter needs a long sequence of flame observations, the assimilation of which requires that multiple instances of the G-equation solver be run for hours.  The Ensemble Kalman Filter can also get trapped in local optima if the ensemble of simulations is initialized poorly. Our current study evaluates the feasibility of performing the same task accurately with a drastically lower compute budget and fewer observations, using an amortized neural network-based inference technique.

\section{Data: experiments and simulations}
\subsection{Experimental observations of ducted premixed Bunsen flame}
\begin{figure}[h]
  \centering
  \includegraphics[width=200 pt]{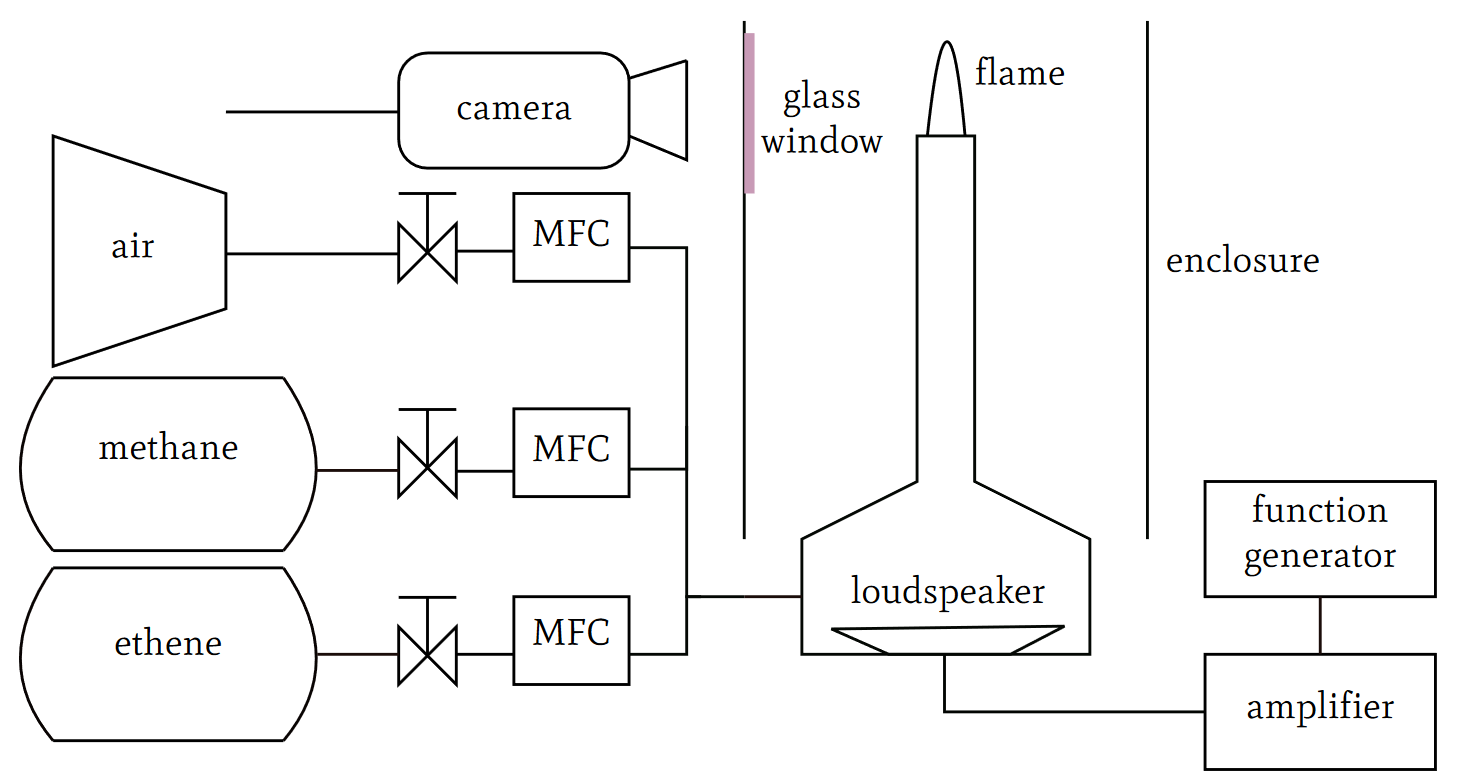}
  \caption{The experimental setup.}
  \label{fig::Apparatus}
\end{figure}
Figure 1 shows a schematic of the experimental setup. It has a premixed, laminar Bunsen flame inside a cylindrical enclosure, with an optical access window for a high-speed camera. A loudspeaker, driven by an  amplified sinusoidal signal, is mounted upstream of the flame for acoustic forcing. Experiments were performed at different fuel compositions (methane: ethene ratios), flow rates, excitation frequencies (250 - 450 Hz) and excitation amplitudes. The flame dynamics are recorded with a Phantom v4.2 CMOS camera at a resolution of 1200x800 pixels and a frame rate $f_s = 2500$ fps. Thresholding is applied to the grayscale frames to detect a thick band around the flame front. For each row of pixels in the image, the mean $x$ coordinate of the pixels within the band, weighted by their intensities, is used to compute a single $x$ coordinate. Cubic splines with 28 knots are then fitted through the $(x,y)$ coordinates to obtain a unique $x$ location of the flame front corresponding to every $y$. The vector of flame front $x$-coordinates could be used directly as input observations for inference. We are interested in modelling the variation of the instantaneous heat release rate and flame areas are a better proxy for heat release rates. We therefore divide the domain vertically into 90 horizontal strips and compute the flame area corresponding to the flame segment it contains. Each frame is thus converted into a 90-dimensional vector of flame areas. 10 successive frames from a video recording are stacked to form an observation $\mathbf{z}_i$, a 900-long vector from which we need to infer the G-equation parameters $\boldsymbol{\theta}_i$ likely to have generated it.

\subsection{Reduced order G-equation model}
The G-equation is a kinematic flame model that models the flame as an infinitely-thin front propagating into unburnt gases at a fixed speed $s_{L}$ \cite{dowling1997nonlinear}. It uses a level-set formulation where the flame front is defined as the $G=0$ contour of the two-dimensional time-varying G-field $G(x,y,t)$. The G-equation is written as 
\begin{equation}
\frac{\partial G}{\partial t} + \mathbf{v}.\nabla G= s_{L} |\nabla G|
\end{equation}
Here, flame speed $s_L = s_{L,u}(1-M_k \kappa)$ is a function of the unstretched flame speed $s_{L,u}$, the flame curvature $\kappa$ and the Markstein length $M_k$. The prescribed velocity field $\mathbf{v} = (U(x) + v')\mathbf{j} + u'\mathbf{i}$ is the sum of a constant parabolic base flow $U(x) = U(1+v_a-2v_ax^2)$ and continuity-obeying velocity perturbations $v'(x,y,t) = U \epsilon_a \mathrm{sin}(St(Ky-t))$ and $v'(x,y,t) = -\frac{U\epsilon_a K St}{\beta}x \mathrm{cos}(St(Ky-t))$. $U$ is a characteristic flow speed, $v_a$ determines the parabolicity of the base flow velocity profile, $\epsilon_a$ is the amplitude and $U/K$ the phase speed of velocity perturbations, $St$ is the Strouhal number and $\beta$ is the aspect ratio of the unforced flame.
\begin{figure}[h]
  \centering
  \includegraphics[width=390 pt]{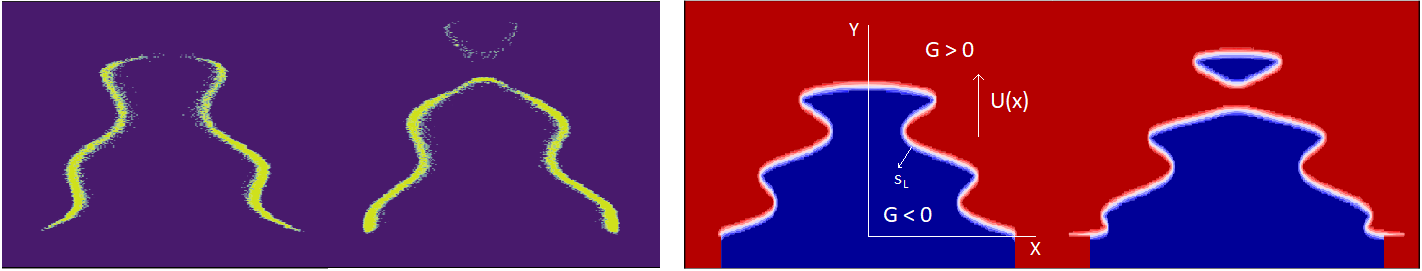}
  \caption{Left: Experimentally observed dynamics of an acoustically excited flame. Right: A flame generated by the G-equation.}
  \label{fig::ExpvsModel}
\end{figure}
Any flame dynamics that can be generated by the G-equation may be uniquely specified by a set of 6 parameters which we need to infer from experimental observations: $K$, $\epsilon_a$, $M_k$, $v_a$, $St$ and $\beta$. To generate our simulation dataset, we sample quasi-randomly from the prior over parameters $P(\boldsymbol{\theta})$, assumed uniform within the hyper-rectangle defined by the following bounds $0 \leq K \leq 2.5$, $0 \leq \epsilon_a \leq 1.0$, $0.02 \leq M_k \leq 0.08$, $0.0 \leq v_a \leq 1.0$, $0.5 \leq St \leq 125.0$, $2.0 \leq \beta \leq 10.0$ and $0.08 \leq f_s \leq 0.20$. For each sampled set of parameters $\boldsymbol{\theta}_i$, the G-equation is solved using the LSGEN2D solver \cite{hemchandra2009dynamics} and the solution is time marched until a limit cycle is reached. The simulated flames then undergo the same data pre-processing steps outlined above for the experimental flames. The $x$-coordinate of the zero level set is identified for each $y$, the flame area vectors are computed for each frame and sequences of 10 frames, each separated from adjacent ones by a phase difference so as to match the experimental frame rate $f_s$, are stacked to form observation vectors $\mathbf{z}_i$ corresponding to $\boldsymbol{\theta}_i$. We compute limit cycle oscillations for $11800$ different $\boldsymbol{\theta}_i$-s, from which we generate 2.4 million simulated observations $\mathbf{z}_i$. $10\%$ of the $\boldsymbol{\theta}_i$-s (240000 $\mathbf{z}_i$-s) are held out in the test set.

\section{Inference using heteroscedastic Bayesian Neural Networks}

We assume that the posterior $p(\boldsymbol{\theta}|\mathbf{z})$ can be modelled as a Bayesian neural network $\boldsymbol{\mu}(\mathbf{z}; \mathbf{w})$ with heteroscedastic Gaussian aleatoric uncertainty $\boldsymbol{\sigma}(\mathbf{z}; \mathbf{w})$. Although the Gaussian assumption may seem restrictive, previous studies as well as our physical intuition indicate that the posterior here is unimodal and weakly correlated. Multi-modality, too, can be dealt with in this framework by using a Bayesian mixture density network instead of a simple neural network, but this has not been considered here. The neural network has a simple MLP architecture of 3 fully connected ReLU layers with 100 units each and a final sigmoid layer that forces the output parameters as well as the predicted aleatoric noise to lie between 0 and 1 (input observations and output parameters are both re-scaled into the range $[0, 1]$). A He normal \cite{7410480} prior is used for the neural network parameters.

The size of the dataset calls for scalable approximate inference techniques for training the neural network. Here we use approximately Bayesian ensembling using randomized maximum a posteriori (MAP) sampling \cite{pearce2018uncertainty}. Ensembling had already been shown \cite{lakshminarayanan2017simple} to be empirically effective at providing calibrated estimates of uncertainty for neural networks and the randomized MAP sampling approach grounds this in Bayesian theory. For the $j$-th neural network ensemble member, we draw a sample from the prior distribution over neural network parameters (assumed multivariate normal) $\mathbf{w}_{anc,j} \sim \mathcal{N}(0, \boldsymbol{\Sigma}_{prior})$ and compute the MAP estimate corresponding to a prior re-centered at $\mathbf{w}_{anc, j}$.  Given $N_{D}$ parameter-observation pairs $\{ \boldsymbol{\theta}_i, \mathbf{z}_i\}$, we minimize the following loss function for the $j$-th neural network
\begin{equation}
\textrm{Loss}_{j} = \sum_{i=1}^{N_{D}}(\boldsymbol{\theta}_i-\boldsymbol{\mu}_j(\mathbf{z}_i))^{T} \boldsymbol{\sigma}_j(\mathbf{z})^{-1} (\boldsymbol{\theta}_i-\boldsymbol{\mu}_j(\mathbf{z}_i)) + \sum_{i=1}^{N_{D}}\textrm{log}(|\boldsymbol{\sigma}_j(\mathbf{z})|) +  \|  \boldsymbol{\Sigma}_{prior}^{-1/2}(\mathbf{w}_j - \mathbf{w}_{anc, j})\|_2^{2}
\end{equation}
The prediction of a trained ensemble with $M$ neural networks is therefore a mixture of $M$ Gaussians, each centered at $\boldsymbol{\mu}_j(\mathbf{z}_i)$ with a variance of $\boldsymbol{\sigma}_j(\mathbf{z})$. For computational convenience, we approximate this mixture as a single Gaussian with mean $\frac{1}{M}\sum_j \boldsymbol{\mu}_j$ and variance $\frac{1}{M}\sum_j \boldsymbol{\sigma}_j + \frac{1}{M}\sum_j \boldsymbol{\mu}^2_j - (\frac{1}{M}\sum_j \boldsymbol{\mu}_j)^2$, following similar treatment in \cite{lakshminarayanan2017simple}. This also allows us to decompose the total predictive uncertainty into an aleatoric component (first term) and an epistemic component (second and third terms).

\section{Results}
Results on the test set of simulated observations (Figure 3) indicate that accurate estimates of G equation parameters are recovered by the neural network. The correlation coefficient $\rho$ between true and predicted parameter values are 0.982, 0.994, 0.971, 0.993, 0.976 and 0.990 for $K$, $\epsilon_a$, $M_k$, $v_a$, $St$ and $\beta$, respectively. Estimates of parameter uncertainty are also well-calibrated, with particularly high uncertainties in $K$ and $St$ predicted for datapoints with low $\epsilon_a$-s (amplitude of perturbation), which is physically sensible, as it is not possible to recover these parameters from an unperturbed flame. Given a 10-frame observation vector as input, the neural network takes \textasciitilde $0.1$ milliseconds on our Tesla K80 GPU to make parameter predictions. Including the pre-processing time for thresholding, spline fitting and area calculation, predictions can therefore be made in less than a millisecond for every few seconds of acquired video data.
\begin{figure}
  \centering
  \includegraphics[width=390 pt]{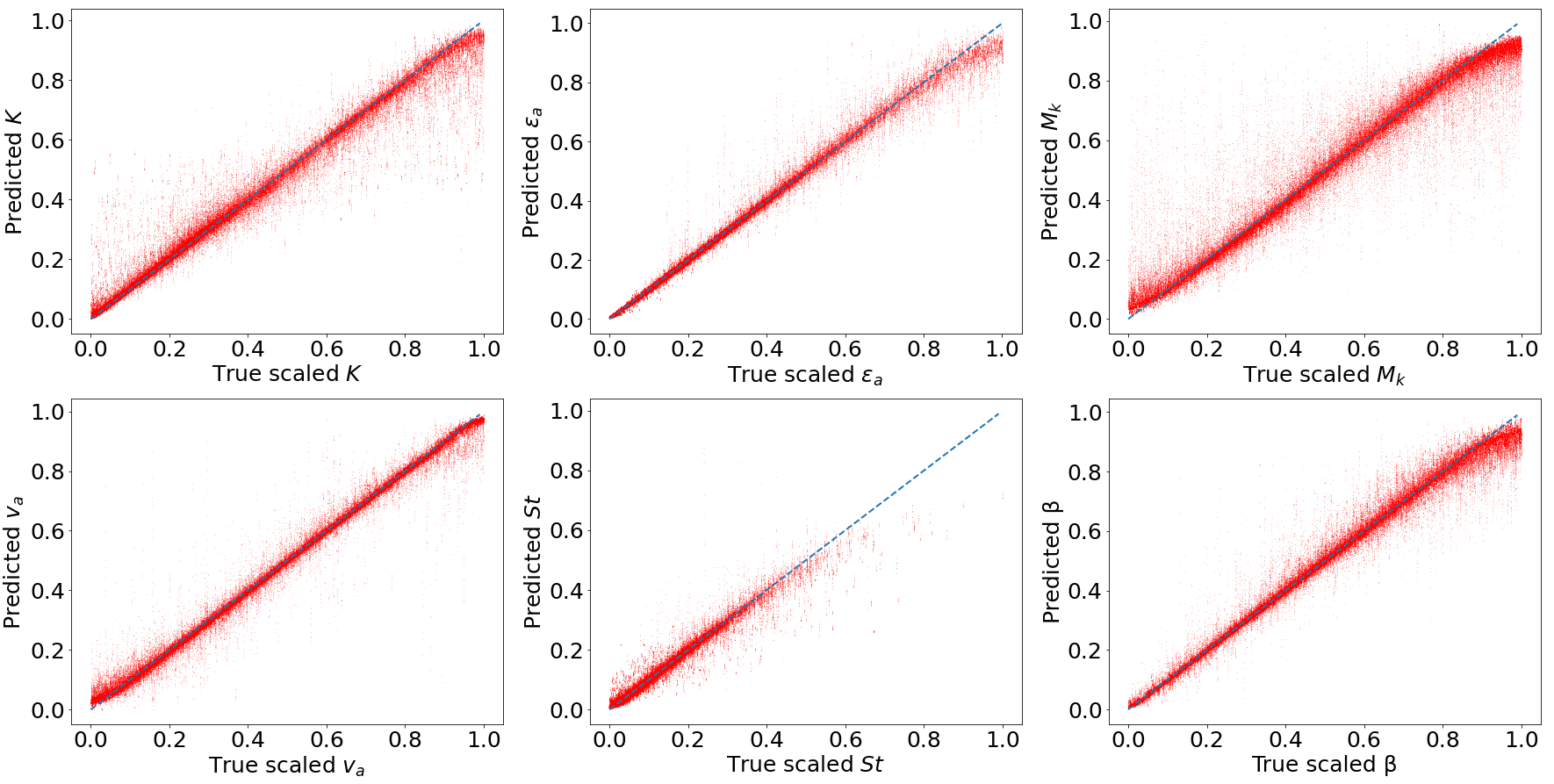}
  \caption{Scatter plots of true parameters (scaled) versus predicted parameter values for simulated test data.}
\end{figure}

We use the trained network to predict parameter values for 10 experimental flame videos and re-simulate the flames using mean predicted parameter estimates. The re-simulated flames match the dynamics of the real flames closely (Figure 4). The root mean squared difference between $x$-coordinates of simulated and real flame fronts, averaged over all $y$ positions and frames, ranges from 0.0171 to 0.068 units (burner radius is 1 units). The mean re-simulation RMSD across all the flame videos is 0.043 units.
\begin{figure}
  \centering
  \includegraphics[width=250 pt]{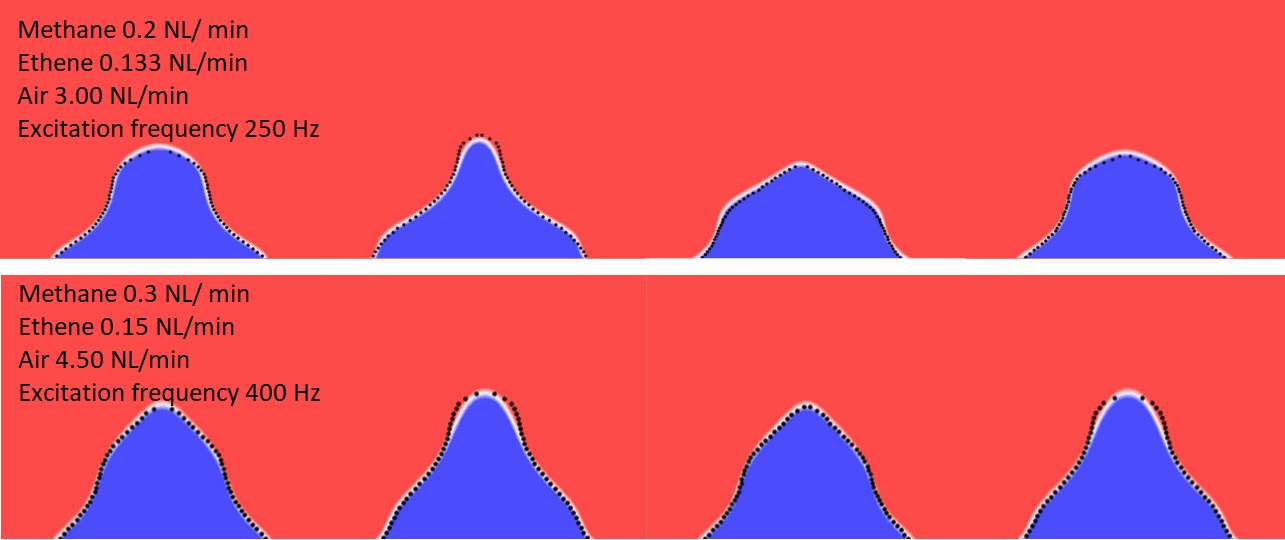}
  \caption{Experimentally observed flames (black dots) overlaid on re-simulated G-equation flames.}
  \label{fig::resimulate}
\end{figure}
\section{Conclusions}
We use a heteroscedastic Bayesian neural network ensemble to calibrate the parameters of the G-equation, a reduced-order model for predicting the dynamics of premixed flames, based on observations. The ensemble was trained on 2.1 million simulated flame observations and results on the test set of simulated flames show that parameters are recovered accurately by the network from only 10 frames of flame data. Reasonable estimates of parameter uncertainties were also obtained. We then use the network to infer parameters from high-speed video footage of acoustically excited Bunsen flames in our lab. Re-simulating the flames using inferred parameters reveals a close agreement between the model output and experimental data, validating our data assimilation. Compared to the Ensemble Kalman filter used for this task in the literature, our neural network based technique can perform assimilate flame videos much more rapidly and obtain accurate parameter estimates using fewer observations.

There are several promising avenues for future work. We will look at assimilating data from more realistic industrial flame configurations as well as integrating our calibrated reduced-order flame models into higher-level thermoacoustic network models of an entire combustor.

\section*{Broader Impact}

This work will lead to better-designed high-energy density combustors, such as those found in rocket and aircraft engines, by enabling designers to assimilate data from experiments more quickly, refine thermoacoustics models more accurately, and therefore design out thermoacoustic oscillations more reliably. 

\begin{ack}
This project has received funding from the European Union’s Horizon 2020 research and innovation program under the Marie Skłodowska- Curie grant agreement number 766264.
\end{ack}

\bibliographystyle{unsrtnat}
\bibliography{bib.bib}

\end{document}